\documentclass[]{TEAI}
\usepackage{helvet}

\usepackage{nicefrac}
\usepackage{xspace}
\usepackage{relsize}
\usepackage{makecell}
\usepackage{bbding}
\usepackage{caption}
\usepackage{float}
\usepackage[export]{adjustbox}

\definecolor{lightblue}{RGB}{220,235,250}
\newcommand{\lightblue}[1]{\cellcolor{lightblue}{#1}}

\newcommand{\system}{\textsc{Ideal}\xspace}

\def\eg{\textit{e.g.}\xspace}

\def\ie{\textit{i.e.}\xspace}

\crefname{section}{Sec.}{Secs.}
\Crefname{section}{Section}{Sections}
\Crefname{table}{Table}{Tables}
\crefname{table}{Tab.}{Tabs.}

\title{\system: In-DEpth ALignment \\ Makes A Discrete Representation AutoEncoder}

\author{
    Yitong Chen\textsuperscript{1,2,*},
    Zijie Diao\textsuperscript{1,*},
    Junke Wang\textsuperscript{1},
    Lingyu Kong\textsuperscript{1},
    Yixuan Ren\textsuperscript{3},
    Bo He\textsuperscript{3},
    Yu-Gang Jiang\textsuperscript{1},
    Zuxuan Wu\textsuperscript{1,2,\textdagger}
}

\affiliation[1]{\mbox{Institute of Trustworthy Embodied AI, Fudan University}}
\affiliation[2]{\mbox{Shanghai Innovation Institute}}
\affiliation[3]{\mbox{University of Maryland, College Park}}

\contribution[*]{Equal contribution}
\contribution[\dagger]{Corresponding author}

\abstract{
Built on pretrained vision foundation models (VFMs), representation autoencoders (RAEs) have recently emerged as a promising approach for constructing semantically rich latent spaces for image generation. However, their reconstruction quality often remains suboptimal, largely because deep VFM representations do not preserve sufficient fine-grained visual detail. This limitation becomes even more severe after discretization, where missing low-level information is difficult to recover. In fact, we observe that shallow VFM features retain considerably richer local appearance and structural detail, which complements the high-level semantics carried by deep features used in existing RAEs. Motivated by this complementary property, we propose \textbf{\system}, an \textbf{\textsc{I}}n-\textbf{\textsc{de}}pth \textbf{\textsc{Al}}ignment framework for discrete representation autoencoding. By jointly aligning quantized tokens with both shallow and deep VFM features, \system enables the resulting discrete visual tokens to preserve both visual fidelity and rich semantics. Extensive experiments demonstrate that \system yields superior reconstruction performance, achieving $\mathbf{0.61}$ rFID on ImageNet and outperforming the previous best method by $\mathbf{0.28}$. When used for autoregressive image generation, \system further produces a gFID of $\mathbf{1.89}$, establishing a new state of the art for autoregressive image generation.

}

\checkdata[Website]{\url{https://github.com/Row11n/IDEAL}}

\begin{document}
\maketitle

\section{Introduction}
\label{sec:intro}

Pretrained vision foundation models (VFMs)\cite{clip,siglip1,siglip2,dinov1,dinov2,dinov3,bolya2026perception} encode images into semantically rich latent spaces that exhibit strong transfer across a broad spectrum of downstream vision tasks. More recently, representation autoencoders (RAEs)\cite{rae} have shown that such frozen VFM features can also serve as effective latent representations for diffusion-based image generation~\cite{stablediffusion,dit,sit}, improving both optimization efficiency and synthesis quality. This emerging connection between representation learning and generative modeling suggests that pretrained representations may offer a strong and scalable foundation for image generation.

However, this promising paradigm still faces a fundamental reconstruction bottleneck. Pretrained VFMs are primarily optimized for semantic discrimination~\cite{siglip2,dinov3}, rather than detail-preserving reconstruction~\cite{vae,vq-vae,wang2024omnitokenizer}. As a consequence, their deep features emphasize high-level semantics but are relatively insensitive to fine-grained visual attributes such as color, texture, and local structure~\cite{svg,dualtoken}. Existing RAEs therefore remain suboptimal for faithful reconstruction, despite their benefits for generation. This issue is further amplified in autoregressive (AR) image generation, where VFM latents must be discretized into visual tokens and missing low-level information is difficult to recover after quantization~\cite{magvitv2,llamagen}.

\begin{figure}[t]
\begin{center}
\centerline{\includegraphics[width=\columnwidth]{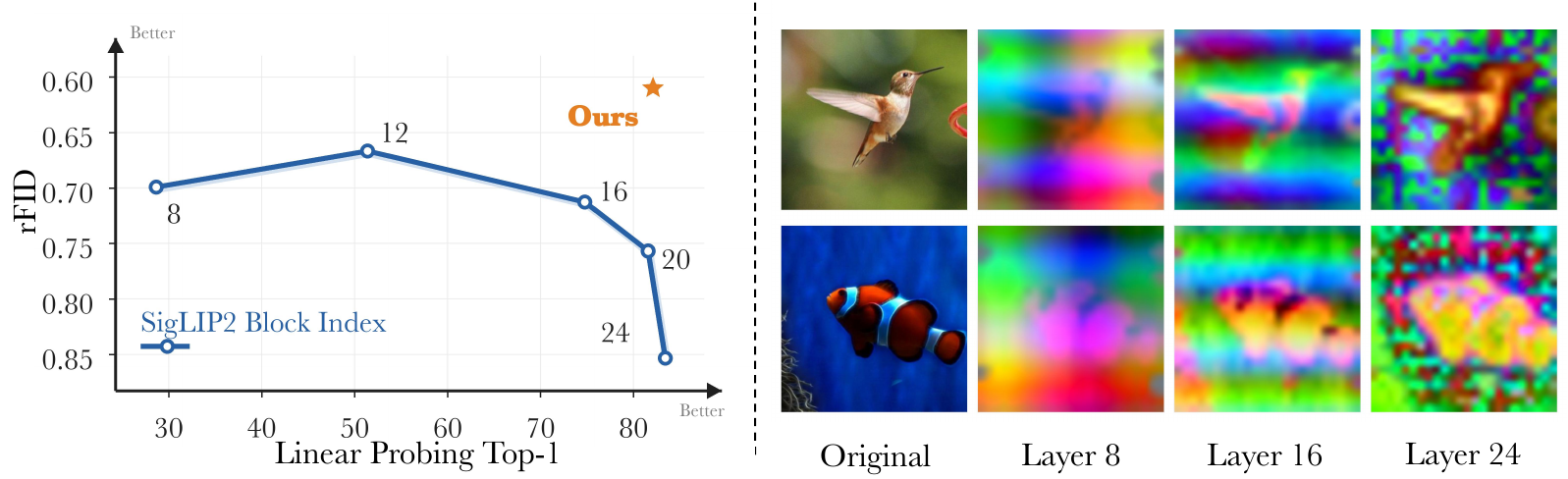}}
\caption{
\textbf{(Left) Depth-wise linear probing of SigLIP2~\cite{siglip2} features.} Each point represents a different VFM block, showing the trade-off between reconstruction fidelity and semantic preservation: shallow blocks reconstruct better but are less semantic, while deeper blocks are more semantic but reconstruct worse. \textbf{(Right) PCA visualization.} By visualizing features across different layers of SigLIP2, we observe a consistent depth-dependent transition: the representations gradually evolve from low-level visual details to high-level semantic concepts.
}
\label{fig:teaser}
\end{center}
\vspace{-0.4in}
\end{figure}

In this work, we ask a simple question: how can discrete representation autoencoding capture fine-grained visual detail without sacrificing high-level semantics? To answer this question, we conduct a systematic depth-wise study by discretizing intermediate VFM representations and evaluating them from two complementary perspectives: semantic preservation and reconstruction fidelity. As shown in~\cref{fig:teaser}, a clear trade-off emerges across layers: shallow representations yield stronger reconstruction but weaker semantics, whereas deeper representations better preserve semantics at the cost of reconstruction fidelity. This trend is consistent with the hierarchical nature of VFMs, whose representations evolve from local texture and geometry in early layers to high-level semantic concepts in later layers~\cite{registers,cambrian,pe}. Taken together, these findings point to a simple yet effective solution: rather than committing to a single layer for tokenization, we enrich deep semantic representations with shallow visual cues, yielding a unified representation that preserves rich semantics while supporting higher-fidelity reconstruction.

With this in mind, we propose \textbf{\system}, a simple yet effective \textbf{\textsc{I}}n-\textbf{\textsc{de}}pth \textbf{\textsc{Al}}ignment framework for discrete representation autoencoding. Rather than choosing a single VFM layer for tokenization, \system combines appearance-rich shallow features with semantically informative deep features prior to vector quantization, forming a unified representation that preserves both visual details and high-level semantics. The resulting tokens are further supervised to recover the corresponding shallow and deep features, explicitly encouraging the discrete representation to retain information from both ends of the hierarchy. Finally, the reconstructed deep features are passed to a lightweight pixel decoder for high-fidelity image reconstruction. In this way, \system turns frozen VFM features into discrete visual tokens that remain both semantically expressive and suitable for faithful reconstruction.

We evaluate \system on ImageNet~\cite{imagenet} from three complementary perspectives: reconstruction fidelity, semantic preservation, and autoregressive generation. For reconstruction, \system obtains an rFID of $\mathbf{0.61}$, outperforming previous tokenizers by $\mathbf{0.28}$ and demonstrating the advantage of incorporating shallow appearance cues. For semantic preservation, the learned discrete representation maintains strong VFM semantics, reaching 80.89\% zero-shot ImageNet classification accuracy. When used for autoregressive image generation, \system yields a gFID of $\mathbf{1.89}$ on ImageNet at $256\times256$ resolution, establishing a new state of the art.

\section{Related Work}

\noindent\textbf{Conventional Tokenizers.}
Existing tokenizers can be roughly divided into two categories: continuous tokenizers and discrete tokenizers. Continuous tokenizers are typically realized as VAEs, with an encoder parameterizing a continuous latent distribution and a decoder reconstructing images from it~\cite{vae,betaVAE,stablediffusion}. In contrast, discrete tokenizers (\eg, VQ-VAE~\cite{vqvae}) learn a finite codebook and quantize encoder features via nearest-neighbor lookup to yield token indices.
Building on VQ-VAE, VQGAN~\cite{vqgan} augments the reconstruction objective with perceptual and adversarial losses, while ViT-VQGAN~\cite{vitvqgan} further modernizes the tokenizer with Transformer-based architectures.
Recent advances refine VQ-based tokenizers along two axes: improved quantization strategy to reduce discretization error~\cite{rqvae,PQ,bsq,fsq,magvitv2,openmagvitv2}, and more stable codebook update approach to mitigate codebook collapse~\cite{vqgan-lc,vqvae2,simvq}.

\noindent\textbf{VFM-based Tokenizers.}
Despite steady progress, most visual tokenizers still lack global semantic structure, which is significant for generation quality~\cite{rae}.
Recent advances show that incorporating pretrained VFM semantics during tokenization~\cite{yao2025vavae} or generation~\cite{repa,reg} can substantially improve generation quality and training efficiency.
These findings have spurred continuous semantic tokenizers like RAE~\cite{rae}, to directly apply tokenization on VFM features. 
FAE~\cite{fae} then successfully reduces the high dimensional latent space of VFMs to a lower dimension using a single attention layer.
On the discrete side, VQRAE~\cite{vqrae} introduces vector quantization into the RAE framework to obtain discrete tokens. VFMTok~\cite{vfmtok} discretizes multi-scale frozen VFM features into codebook indices with deformable attention layers~\cite{DETR}.
DINO-Tok~\cite{dinotok} stabilizes vector quantization in DINO~\cite{dinov2,dinov3} latent space through global PCA reweighting.

\noindent\textbf{Autoregressive Visual Generation.}
With a strong discrete visual tokenizer, images and videos can be compressed into discrete sequences suitable for next-token prediction.
Autoregressive models then perform sequen\-ce modeling over these tokens and generate diverse high-quality images~\cite{llamagen,parti,wang2025simplear,wang2026omnigen} and videos~\mbox{\cite{tats,videogpt}}.
VAR~\cite{var} further redefines autoregressive learning from raster-scan next-token prediction to coarse-to-fine next-scale prediction. xAR~\cite{xar} extends the autoregressive framework further by introducing next-X prediction, enabling flexible prediction targets such as tokens, cells, 
subsamples, and entire images.
\section{Method}

\subsection{Preliminary: Vector Quantized Image Tokenizers}
\label{sec:prelim_vq}
A quantized image tokenizer is commonly formulated as an encoder $E(\cdot)$, a vector-quantizer $\mathrm{VQ}(\cdot)$ with a learnable codebook $C(\cdot)$, and a decoder $D(\cdot)$.
Given an input image $x \in \mathbb{R}^{H \times W \times 3}$, the encoder first compresses it into a 2D patch embedding, and then applies a CNN/ViT backbone to produce the latent embedding $z$.
\begin{equation}
z = E(x) \in \mathbb{R}^{H/p \times W/p \times d},
\label{eq:encode}
\end{equation}
where $p$ denotes the downsampling patch size and $d$ is the channel dimension.
The quantizer maintains a codebook $C = \{c_k\}_{k=1}^{K}$ with each $c_k \in \mathbb{R}^{d}$.
For each spatial location $i$, the continuous embedding $z_i$ is mapped to its nearest codebook entry:
\begin{equation}
\mathrm{VQ}(z_i) = \tilde{z}_i = c_{k_i},
\quad
k_i = \arg\min_{k \in \{1,\dots,K\}} \lVert z_i - c_k \rVert_2 .
\label{eq:vq_assign}
\end{equation}
The resulting discrete representation is the index map $\{k_i\}$, which can be flattened into a token sequence for AR modeling.

De-quantization retrieves the corresponding embeddings $\tilde{z}$ from the indices and decodes them back to the image domain. In practice, the decoder often consists of a feature-decoding backbone followed by a lightweight pixel head.
\begin{equation}
\hat{x} = D(\tilde{z}) = D(\mathrm{VQ}(z)) .
\label{eq:ae_recon}
\end{equation}
To optimize the codebook, we use the standard VQ objective
\begin{equation}
\mathcal{L}_{\mathrm{VQ}}
=
\sum_i \left\lVert \mathrm{sg}(z_i) - c_{k_i} \right\rVert_2^2
+
\beta \left\lVert \mathrm{sg}(c_{k_i}) - z_i \right\rVert_2^2,
\label{eq:vq_loss}
\end{equation}
where $\mathrm{sg}(\cdot)$ denotes the stop-gradient operator~\cite{stopgradient} and $\beta$ is the weight of commitment loss~\cite{vqvae}.

For image reconstruction, we minimize an auto-encoding loss
\begin{equation}
\mathcal{L}_{\mathrm{AE}}
=
\mathcal{L}_2(x,\hat{x})
+
\mathcal{L}_{\mathrm{P}}(x,\hat{x})
+
\lambda_{\mathrm{G}}\,\mathcal{L}_{\mathrm{G}}(\hat{x}),
\label{eq:ae_loss}
\end{equation}
where $\mathcal{L}_2$ is a pixel-wise reconstruction loss, $\mathcal{L}_{\mathrm{P}}$ is a perceptual loss (\eg, LPIPS~\cite{LPIPS}), and $\mathcal{L}_{\mathrm{G}}$ is an adversarial loss (\eg, PatchGAN~\cite{PATCHGAN}) weighted by $\lambda_{\mathrm{G}}$.

In this work, we follow the quantized-tokenizer paradigm above and focus on learning discrete codes that are suitable for AR modeling while preserving VFM semantics.

\begin{figure*}[t]
\begin{center}
\centerline{\includegraphics[width=\textwidth]{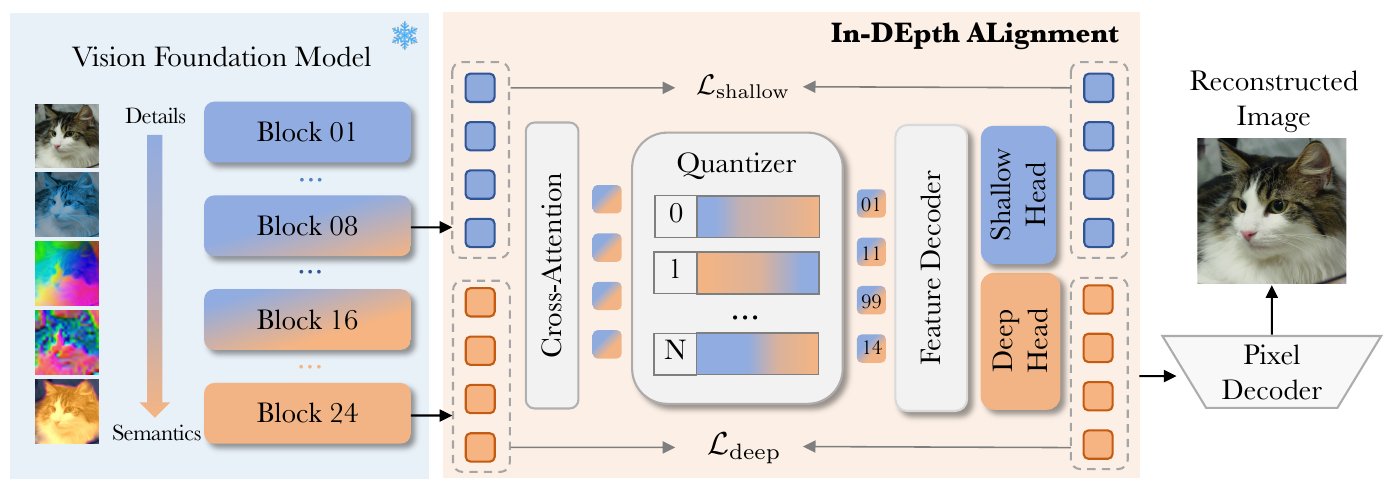}}
\caption{
 \textbf{Illustration of \system.} \system first extract shallow and deep features from a frozen VFM. A lightweight cross-attention module then fuses them into a unified representation. After vector quantization, a feature decoder reconstructs both shallow and deep features. The reconstructed deep semantic feature is finally mapped to pixels by a lightweight pixel decoder for image reconstruction.
}
\label{Fig2}
\end{center}
\end{figure*}

\subsection{Semantic-Spatial Complementarity in VFMs}
\label{sec:pilot_vfm}

\noindent
\begin{minipage}[c]{0.55\textwidth}
\paragraph{Protocol.}
To understand which VFM features can provide fine-grained details for discrete semantic tokenization, we conduct a depth-wise probe by freezing a pretrained VFM $\Phi(\cdot)$ and tokenizing its intermediate features, as mentioned in~\cref{sec:intro}.
Given an image $x$, we extract a layer feature $f^{(\ell)}=\Phi_{\ell}(x)$, quantize it with the VQ module in~\cref{sec:prelim_vq}, and reconstruct it in two steps: a feature decoder produces a reconstructed feature, which is then mapped to pixels by a decoder.
We evaluate each layer $\ell$ using (i) pixel reconstruction FID after the decoder and (ii) linear probing classification Top-1 accuracy on
the reconstructed feature.
\paragraph{Layer-wise trade-off.}
We probe a set of VFM layers $\ell \in \{8,12,16,20,24\}$ as tokenization targets. As shown in Table~\ref{tab:vfm_layer_probe}, shallow-layer features are easier to reconstruct with 
\end{minipage}
\hfill
\begin{minipage}[c]{0.4\textwidth}
    \label{tab:vfm_layer_probe}
    \centering
    \captionsetup{type=table}
    \caption{\textbf{Layer-wise probing results on SigLIPv2~\cite{siglip2} features.} We report reconstruction fidelity using rFID and semantic preservation using linear probing classification Top-1 accuracy. Deeper layers retain more semantics, but leads to inferior reconstruction performance.}
    \begin{tabular}{lcc}
    \toprule
    Layer & rFID$\downarrow$ & LP-Top1$\uparrow$ \\
    \midrule
    8  & \underline{0.69} & 28.66  \\
    12 & \textbf{0.66} & 51.40  \\
    16 & 0.71 & 74.78 \\
    20 & 0.75 & \uline{81.57} \\
    24 & 0.85 & \textbf{83.43} \\
    \bottomrule
    \end{tabular}
\vspace{0.15in}
\end{minipage}
higher pixel fidelity, while their reconstructed features exhibit weak semantic transfer. Meanwhile, deeper-layer features preserve semantic ability better after quantization, but their pixel reconstruction performance tend to degrade. 
Overall, VFMs provide complementary signals across depth: shallow features are more reconstruction-friendly, whereas deep features are more semantic.

% \begin{table}[t]
% \centering
% \setlength{\tabcolsep}{6pt}
% \renewcommand{\arraystretch}{1.15}
% \caption{Layer-wise probing results on discretizing intermediate SigLIPv2~\cite{siglip2} features. We report reconstruction fidelity using rFID and semantic preservation using Zero-Shot classification Top-1 accuracy. Deeper layers retain more semantics, but leads to inferior reconstruction performance.}
% \begin{tabular}{lcc}
% \toprule
% VFM block & rFID$\downarrow$ & ZS-Top1$\uparrow$ \\
% \midrule
% 8  & \underline{0.69} & 0.25  \\
% 12 & \textbf{0.66} & 3.45  \\
% 16 & 0.71 & 27.19 \\
% 20 & 0.75 & 66.74 \\
% 24 & 0.85 & \underline{81.52} \\
% SigLIPv2 & N/A & \textbf{83.23} \\
% \bottomrule
% \end{tabular}
% \label{tab:vfm_layer_probe}
% \end{table}

\subsection{\system}
\label{sec:semtok}

Motivated by the complementary behavior of shallow and deep VFM features, we propose \textbf{\system}, a VFM-based semantic tokenizer that produces discrete token indices for AR modeling and preserves semantic capability after de-quantization. The overall architecture of \system is illustrated in Figure~\ref{Fig2}.
\paragraph{Frozen VFM encoder and fusion before quantization.}
We freeze a pretrained VFM $\Phi(\cdot)$ as the encoder.
Given an image $x$, we extract a shallow feature $f^{(s)}=\Phi_{\ell_s}(x)$ and a deep feature $f^{(d)}=\Phi_{\ell_d}(x)$ from two VFM layers.
In our setting, both features are sequences with matched shapes, \ie, $f^{(s)}, f^{(d)} \in \mathbb{R}^{B \times L \times D}$, allowing fusion without any additional resizing or projection.
We implement $\mathrm{AttnFuse}(\cdot)$ as a single lightweight cross-attention block where deep features provide queries and shallow features provide keys/values, followed by a Feed Forward Network(FFN) to produce the fused representation $z$.

\begin{equation}
z =\mathrm{AttnFuse}\!\left(f^{(d)},\,f^{(s)}\right).
\label{eq:vfm_fuse}
\end{equation}

We adopt the VFM’s original normalization for $f^{(d)}$ and a learnable normalization for $f^{(s)}$.

\paragraph{Vector quantization.}
To avoid introducing additional complexity, we quantize $z$ using the standard VQ formulation in Equation~\ref{eq:vq_assign}, yielding discrete token indices $y$ and de-quantized embeddings $\tilde{z}$.
We apply an $\ell_2$ normalization on codebook vectors to stabilize nearest-neighbor assignment during training.
Following common practice~\cite{vitvqgan}, we apply down-factorization to map the fused feature $z$ into a lower-dimensional quantization space before lookup, and recover the original dimension after de-quantization.
This design mitigates codebook collapse and achieves full codebook utilization in our experiments.
The resulting token indices $y$ are used for AR modeling in \cref{sec:ar_generation}.
\paragraph{Two-step decoding with dual feature heads.}
We decode $\tilde{z}$ using a ViT backbone feature decoder $D_{\mathrm{feat}}$ 
to reconstruct the unified feature.
\begin{equation}
g = D_{\mathrm{feat}}(\tilde{z}).
\label{eq:feat_decode}
\end{equation}
Following previous work~\cite{vfmtok}, we also append a [CLS] token and several register tokens to the input sequence to enhance representation learning and capture global context.
These tokens are not used for reconstruction.

From $g$, we apply two lightweight linear heads to reconstruct the deep semantic feature and the shallow spatial feature, producing $\hat{f}^{(d)}$ and $\hat{f}^{(s)}$, respectively.
We use $\hat{f}^{(d)}$ as the interface feature for semantic preservation evaluation, and also feed it into the pixel decoder to reconstruct the image:
\begin{equation}
\hat{x} = D_{\mathrm{pixel}}\!\left(\hat{f}^{(d)}\right).
\label{eq:pixel_decode}
\end{equation}
\paragraph{Objectives}
In addition to the standard VQ loss $\mathcal{L}_{\mathrm{VQ}}$ (Equation~\ref{eq:vq_loss}) and the auto-encoding loss $\mathcal{L}_{\mathrm{AE}}$ (Equation~\ref{eq:ae_loss}), we align reconstructed features with their VFM targets on both the deep and shallow branches. These alignment terms encourage the feature decoder to produce representations that preserve both semantic structure and fine-grained details.

The deep alignment loss is
\begin{equation}
\mathcal{L}_{\mathrm{deep}}
=
\left\lVert \hat{f}^{(d)} - f^{(d)} \right\rVert_2^2
+
\left(1-\cos\!\left(\hat{f}^{(d)}, f^{(d)}\right)\right),
\label{eq:sem_align}
\end{equation}
and the shallow alignment loss is
\begin{equation}
\mathcal{L}_{\mathrm{shallow}}
=
\left\lVert \hat{f}^{(s)} - f^{(s)} \right\rVert_2^2
+
\left(1-\cos\!\left(\hat{f}^{(s)}, f^{(s)}\right)\right).
\label{eq:spa_align}
\end{equation}
For the adversarial term in $\mathcal{L}_{\mathrm{AE}}$, we replace the conventional PatchGAN discriminator ~\cite{PATCHGAN} with a frozen DINOv1-s model ~\cite{dinov1}, yielding semantically meaningful adversarial guidance that consistently improves reconstruction quality.
The full objective is then
\begin{equation}
\mathcal{L}
=
\mathcal{L}_{\mathrm{AE}}
+
\mathcal{L}_{\mathrm{VQ}}
+
\mathcal{L}_{\mathrm{deep}}
+
\mathcal{L}_{\mathrm{shallow}}.
\label{eq:full_loss}
\end{equation}

\subsection{Autoregressive Image Generation}
\label{sec:ar_generation}

Once a tokenizer is trained, its discrete codes can be modeled by an autoregressive Transformer via next-token prediction.
Let $y = (y_1,\dots,y_T)$ denote the flattened token indices, and let $c$ be the conditioning signal such as a class label or text embedding.
An AR model parameterized by $\theta$ factorizes the likelihood as
\begin{equation}
p_\theta(y \mid c) = \prod_{t=1}^{T} p_\theta(y_t \mid y_{<t}, c),
\end{equation}
and is trained with the standard cross-entropy objective
\begin{equation}
\mathcal{L}_{\mathrm{AR}} = - \sum_{t=1}^{T} \log p_\theta(y_t \mid y_{<t}, c).
\label{eq:ar_loss}
\end{equation}
During sampling, the model generates $\hat{y}$ sequentially, after which the tokenizer decoder maps $\hat{y}$ back to an image.
In the AR model, we use 2D RoPE~\cite{rope} to better capture spatial locality.  

\section{Experiments}
\label{sec:exp}

\subsection{Setup}
\label{sec:exp_setup}

\paragraph{Image tokenizer.}
We train \system on ImageNet-1K ~\cite{imagenet} and report results on the validation set.
Unless stated otherwise, we follow the standard tokenizer training protocol in VQGAN~\cite{vqgan} to ensure fair comparison.
Since many VFMs are pretrained with an input resolution of $384{\times}384$, we train the tokenizer on images resized to the same resolution.
We adopt SigLIP2-Large-384~\cite{siglip2} as the frozen VFM encoder and use features from the $8$th and $24$th Transformer(deepest) blocks as $f^{(s)}$ and $f^{(d)}$ respectively.
The feature decoder is a $6$-layer Transformer, consistent with prior work ~\cite{vfmtok}.

For reporting reconstruction metrics, we resize reconstructed images to $256{\times}256$, matching the evaluation protocol in ~\cite{llamagen}.
We use a VQ codebook with size $K{=}16384$ and vector dim $d{=}64$.

\paragraph{Class-conditional autoregressive generation.}
We evaluate class-conditional AR generation by training AR models on the discrete token sequences produced by our tokenizer.
We evaluate \system with class-conditional autoregressive (AR) generation on ImageNet-1K at $256{\times}256$.
Following LlamaGen recipe~\cite{llamagen}, samples generated by AR models are $384{\times}384$ and are resized to $256{\times}256$ for metric computation.
We train four AR model variants at different scales: Base (111M), Large (343M), XXL (1.4B), and 3B parameters. Models with fewer than $1$B parameters are trained for $300$ epochs, and larger models are trained for $200$ epochs.  

\paragraph{Evaluation metrics.}
For tokenizer reconstruction, we report reconstruction Fr\'echet Inception Distance ~\cite{FID} (rFID) and reconstruction Inception Score ~\cite{IS} (rIS) as main metrics.
For generation quality, we use generation Fr\'echet Inception Distance (gFID) and generation Inception Score (gIS) as primary metrics. We additionally report sFID, Precision, and Recall ~\cite{recall} for completeness.

To quantify how well our tokenizer preserves VFM semantics, we report zero-shot ImageNet-1K classification accuracy (ZS Top-1/Top-5) following CLIP~\cite{clip}.
\subsection{Main Results}

\noindent
\begin{minipage}[c]{0.38\textwidth}
\paragraph{Image reconstruction.}
We compare \system with representative discrete image tokenizers, including conventional visual tokenizers like VQGAN~\cite{vqgan} and semantic tokenizers like VFMTok~\cite{vfmtok}.
As shown in \cref{tab:tokenizer_rfid_ris}, \system achieves $0.61$ rFID, outperforming prior VQ-based baselines under comparable settings while maintaining $100\%$ codebook utilization.
Beyond pixel fidelity, \system also attains the highest rIS of $230.4$, indicating \system's strong semantic consistency between reconstructed and original images.
This suggests that \system improves reconstruction without sacrificing the semantic structure inherited from the VFM.
Refer to \cref{tab:tokenizer_comparison_256} for a fully controlled comparison on 256 resolution.

Overall, \system can achieve superior performance in both reconstruction fidelity and semantic consistency while still maintaining 100\% usage, demonstrating that our design can substantially improve discrete autoencoding.
\end{minipage}
\hfill
\begin{minipage}[c]{0.60\textwidth}
    \centering
    \captionsetup{type=table}
    \small
\setlength{\tabcolsep}{1.1pt}
\renewcommand{\arraystretch}{1}
\caption{\textbf{System-level reconstruction performance and codebook utilization.}
`$f$' denotes the downsampling ratio, `Size' the codebook size, `Dim.' the codebook vector dimension, and `\#Res.' the tokenization resolution. Results with resolution higher than 256 are resized to 256 when computing the metrics.
$^{\mathrm{oim}}$ indicates tokenizers trained on OpenImages~~\cite{openimage}.}
\begin{tabular}{l|ccccccc}
\toprule
Method & $f$ & Size & Dim. & \#Res. & rFID$\downarrow$ & rIS$\uparrow$ & Usage (\%) \\
\midrule
\multicolumn{8}{l}{\textit{\textbf{Conventional tokenizer}}} \\
TiTok~~\cite{titok}                & -- & 8192  & 64  & 256  & 1.05 & 191.5 & 100   \\
ImageFolder~~\cite{imagefolder}    & -- & 32768 & 32  & 256  & 0.69 & 201.5 & 100  \\
VQGAN~~\cite{vqgan}                 & -- & 16384  & 256 & 256 & 4.98 & --    & --   \\
VQGAN~~\cite{vqgan}                 & -- & 8192  & 256 & 256 & 1.49 & --    & --   \\
VQGAN$^{\mathrm{oim}}$~~\cite{vqgan} & -- & 16384 & 4   & 256   & 1.19 & --    & --    \\
ViT\mbox{-}VQGAN~~\cite{vitvqgan}   & -- & 8192  & 32  & 256   & 1.28 & 192.3  & 95.0 \\
% \cmidrule{1-9}
MaskGiT~~\cite{maskgit}            & 16 & --    & --  & 256   & 2.28 & --    & --   \\
VAR~~\cite{var}                    & 16 & 4096  & 32  & 256  & 0.92 & 196.0 & 100 \\
RQ\mbox{-}VAE~~\cite{rqvae}        & 32 & 16384 & 256 & 256 & 1.83 & --    & --  \\
LlamaGen~~\cite{llamagen}             & 16 & 16384 & 8  & 336  & 1.21 & 189.1   & 99.2 \\
LlamaGen~~\cite{llamagen}          & 16 & 16384 & 8   & 384  & 0.95 & 197.3 &  99.7 \\
\midrule
\multicolumn{8}{l}{\textit{\textbf{VFM-based tokenizer}}} \\
% \hdashline
VQRAE~~\cite{vqrae}              & 16 & 16384 & 1536  & 256  & 1.31 & -- & --   \\
DINO\mbox{-}Tok~~\cite{dinotok}   & 16 & 16384$\times$2 & 832  & 256  & 1.15 & -- & --  \\
VFMTok~~\cite{vfmtok}              & -- & 16384 & 12  & 336  & 0.89 & 215.4 & 100   \\
\lightblue{\textbf{\system \emph{(Ours)}}}     & \lightblue{16} & \lightblue{16384} & \lightblue{64}  & \lightblue{384}  & \lightblue{\textbf{0.61}} & \lightblue{\textbf{230.4}} & \lightblue{100}\\
\bottomrule
\end{tabular}
\label{tab:tokenizer_rfid_ris}

\end{minipage}

\begin{table*}[t]
\centering

\end{table*}

\begin{figure}[bt]
\vskip 0.2in
\begin{center}
\centerline{\includegraphics[width=\columnwidth]{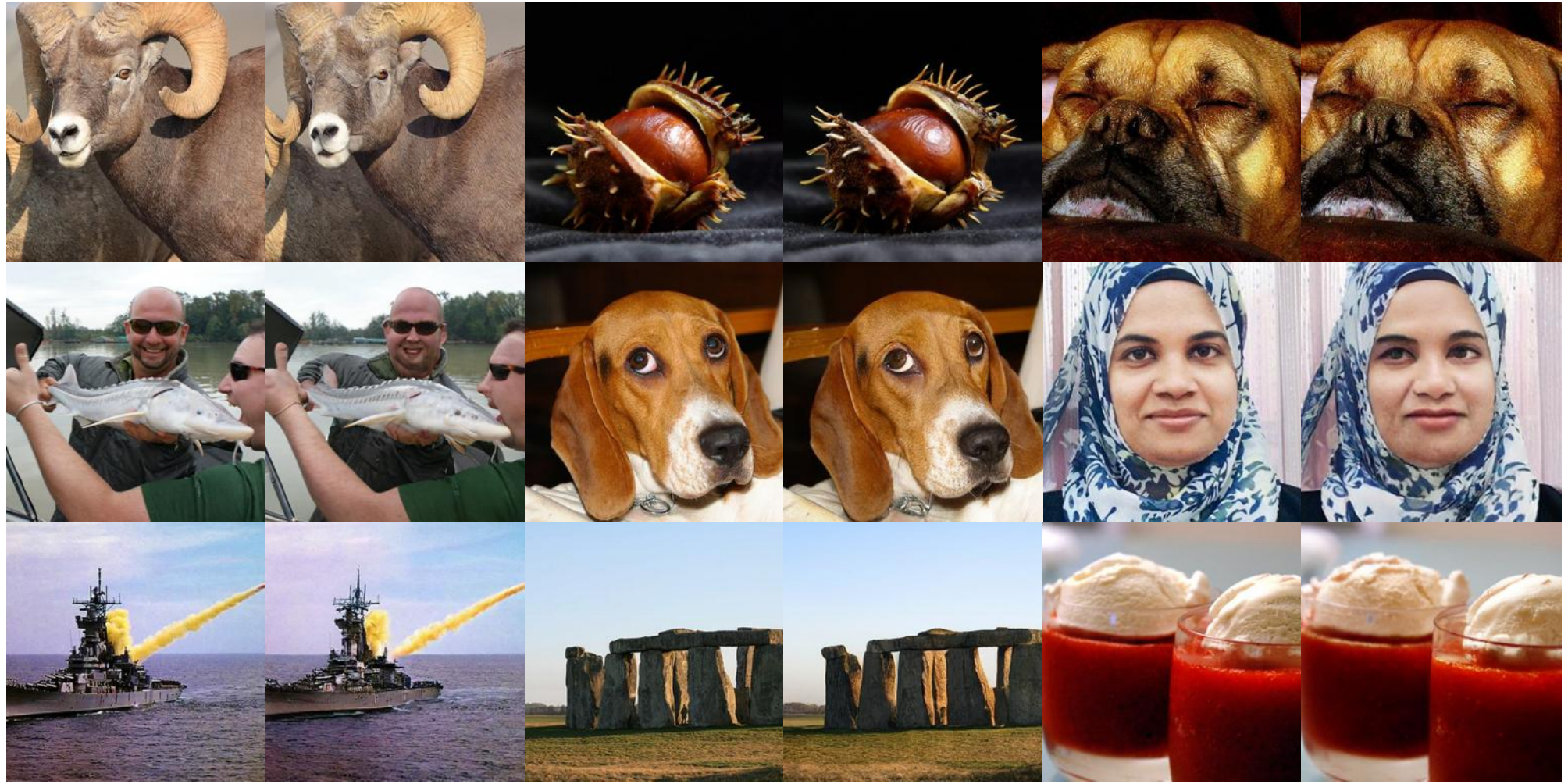}}
\caption{
Visualization of reconstruction results from \system. Left: input image; Right: output image.
}
\label{Fig3}
\end{center}
\end{figure}

\noindent
\begin{minipage}[c]{0.43\textwidth}
\paragraph{Semantic Preservation.}
\label{sec:semantic_preservation}
A primary goal of \system is to preserve the semantic structure of the underlying VFM after discretization and decoding.
We compare our model's performance with the underlying VFM SigLIPv2~\cite{siglip2} on zero-shot ImageNet-1K classification. \cref{tab:zeroshot_imagenet} shows that \system's decoded interface feature can achieve $\mathbf{80.89\%}$ Top-1 and $\mathbf{96.40\%}$ Top-5 accuracy, closely matching SigLIPv2's deepest feature  ($\mathbf{83.23\%}$ vs $\mathbf{97.11\%}$). This indicates that, 
\end{minipage}
\hfill
\begin{minipage}[c]{0.55\textwidth}
    \centering
    \captionsetup{type=table}
    \small
    \setlength{\tabcolsep}{6pt}
    \renewcommand{\arraystretch}{1.15}
    \caption{Zero-shot ImageNet-1K classification accuracy for SigLIPv2~\cite{siglip2} and \system. N/A indicates visual tokenziers do not support zero-shot evaluation. }
    \begin{tabular}{lcc}
    \toprule
    Model / Feature & Top-1 (\%) $\uparrow$ & Top-5 (\%) $\uparrow$ \\
    \midrule
    Conventional tokenizers & N/A & N/A \\
    SigLIP2  & 83.23 & 97.11 \\
    \system  & 80.89 & 96.40 \\
    \bottomrule
    \end{tabular}
    \label{tab:zeroshot_imagenet}
\end{minipage}
despite vector quantization and reconstruction oriented training objectives, the feature reconstructed by decoder can still retain near original VFM semantic structure.

Since we preserve a SigLIPv2-native semantic space, our decoded features naturally remain compatible with SigLIPv2 text embeddings without additional vision--language contrastive training~\cite{clip,unitok}.
This text-interactive property is largely absent in most prior tokenizers, as their decoded features are not compatible with text embeddings and therefore do not support CLIP-style zero-shot classification.

We further evaluate the decoded features on multimodal understanding benchmarks under common used setting~\cite{comp}: the vision encoder is frozen, a newly initialized adapter connects it to LLaMA~3.0 8B, the adapter and LLM are jointly tuned on LLaVA SFT data for one epoch.

\begin{table*}[!ht]
\centering
\small
\setlength{\tabcolsep}{5pt}
\renewcommand{\arraystretch}{1.15}
\caption{Multimodal understanding results.}
\begin{tabular*}{\textwidth}{@{\extracolsep{\fill}}lccccccc}
\toprule
Model &Token & RealWorldQA~\cite{realworldqa} & ChartQA~\cite{chartqa} & OKVQA~\cite{okvqa} & InfoVQA~\cite{doc} & SEED~\cite{seed-bench} & MME~\cite{mme} \\
\midrule
DINOv2 & 576 & 46.26 & 10.80 & 54.12 & 21.33 & 57.00 & 1345 \\
SigLIP2 & 576 & 47.19 & \textbf{13.80} & 59.88 & 20.56 & 58.24 & 1730 \\
\system & 576 & \textbf{52.68} & 12.48 & \textbf{61.06} & \textbf{22.88} & \textbf{68.02} & \textbf{1878} \\
\bottomrule
\end{tabular*}
\label{tab:vqa_semantic_preservation}
\end{table*}

\begin{table*}[t]
\centering
\small
\setlength{\tabcolsep}{5pt}
\renewcommand{\arraystretch}{1.1}
\caption{Class-conditional ImageNet 256 $\times$ 256 generation results with classifier-free guidance (CFG). $\dagger$ indicates re-implementation by ~\cite{vfmtok}; `-re' denotes rejection sampling. Images generated at resolution higher than 256 will be resized to 256 during evaluation. \system-B performs best with a CFG scale of 1.75, while other variants perform best with a CFG scale of 1.25.}
\begin{tabular*}{\textwidth}{@{\extracolsep{\fill}}c|l|ccr|ccccc}
\toprule
\multirow{2}{*}{Type} &
\multirow{2}{*}{Method} &
\multirow{2}{*}{\#Epoch} &
\multirow{2}{*}{\#Params.} &
\multirow{2}{*}{Res.} &
\multicolumn{5}{c}{Generation w/ CFG} \\
\cline{6-10}
& & & & &
gFID$\downarrow$ & sFID$\downarrow$ & gIS$\uparrow$ & Pre.$\uparrow$ & Rec.$\uparrow$ \\
\midrule

% ---------------- Diffusion ----------------
\multirow{4}{*}{Diff.} &
MaskDiT~\cite{maskdit} & 1600 & 675M & 256  & 2.28 & 5.67 & 276.6 & 0.80 & 0.61 \\
& DiT~\cite{dit}       & 1600 & 675M & 256  & 2.27 & 4.60 & 278.2 & 0.83 & 0.57 \\
& SiT~\cite{sit}       & 1600 & 675M & 256  & 2.06 & 4.50 & 270.3 & 0.82 & 0.59 \\
& FasterDiT~\cite{fasterdit} & 400 & 675M & 256 & 2.03 & 4.63 & 264.0 & 0.81 & 0.60 \\
\midrule

% ---------------- Masked modeling ----------------
\multirow{1}{*}{Mask.} &
MaskGiT-re~~\cite{maskgit} & 555  & 227M  & 256 & 4.02 & --   & 355.6& --   & -- \\
\midrule

\multirow{1}{*}{VAR} &
VAR~~\cite{var}        & 350 & 310M & 256 & 3.30 & --   & 274.4 & 0.84 & 0.51 \\
\midrule

% ---------------- Autoregressive ----------------
\multirow{16}{*}[-6.4ex]{AR} &

% -------- Base (≈111M) --------
 \multicolumn{9}{l}{\textbf{Base ($\approx$111M params)}} \\
& TiTok-B$^\dagger$~~\cite{titok}     & 300 & 111M & -- & 6.76 & 7.82 & 175.3 & 0.85 & 0.43 \\
& LlamaGen-B~~\cite{llamagen}         & 300 & 111M & 384 & 6.09 & 7.24 & 182.5 & 0.85 & 0.42 \\
& VFMTok-B~~\cite{vfmtok}             & 300 & 111M & 336 & 3.43 & 5.88 & \textbf{252.2} & \textbf{0.85} & \textbf{0.53} \\
& \lightblue{\textbf{\system-B (Ours)}} & \lightblue{300} & \lightblue{111M} & \lightblue{384} &
\lightblue{\textbf{3.38}} & \lightblue{\textbf{5.18}} & \lightblue{219.8} & \lightblue{0.84} & \lightblue{0.51} \\
\addlinespace[2pt]
\cmidrule(lr){2-10}

% -------- Large (≈343M) --------
& \multicolumn{9}{l}{\textbf{Large ($\approx$343M params)}} \\
& TiTok-L$^\dagger$~~\cite{titok}     & 300 & 343M & -- & 4.03 & 6.93 & 219.5 & 0.84 & 0.52 \\
& LlamaGen-L~~\cite{llamagen}         & 300 & 343M & 384 & 3.07 & 6.09 & 256.1 & 0.83 & 0.52 \\
& VFMTok-L~~\cite{vfmtok}             & 300 & 343M & 336 & 2.75 & 5.58 & \textbf{278.8} & \textbf{0.84} & 0.57 \\
& \lightblue{\textbf{\system-L (Ours)}} & \lightblue{300} & \lightblue{343M} & \lightblue{384} &
\lightblue{\textbf{2.26}} & \lightblue{\textbf{5.10}} & \lightblue{219.71} & \lightblue{0.81} & \lightblue{\textbf{0.58}} \\
\addlinespace[2pt]
\cmidrule(lr){2-10}

% -------- XXL (≈1.4B) --------
& \multicolumn{9}{l}{\textbf{XXL ($\approx$1.4B params)}} \\
& LlamaGen-XXL~~\cite{llamagen}       & 200 & 1.4B & 384 & 2.34 & 6.00 & 253.9 & 0.81 & 0.60 \\
& VFMTok-XXL~~\cite{vfmtok}           & 200 & 1.4B & 336 & 2.19 & 5.53 & \textbf{278.0} & 0.83 & \textbf{0.60} \\
& \lightblue{\textbf{\system-XXL (Ours)}} & \lightblue{200} & \lightblue{1.4B} & \lightblue{384} &
\lightblue{\textbf{1.95}} & \lightblue{\textbf{4.81}} & \lightblue{260.2} & \lightblue{\textbf{0.83}} & \lightblue{0.59} \\
\addlinespace[2pt]
\cmidrule(lr){2-10}

% -------- 3B --------
& \multicolumn{9}{l}{\textbf{3B params}} \\
& LlamaGen-3B~~\cite{llamagen}        & 200 & 3.1B & 384 & 2.19 & 5.97 & 263.3 & 0.82 & 0.58 \\
& VFMTok-3B~~\cite{vfmtok}            & 200 & 3.1B & 336 & 2.07 & 6.23 & \textbf{280.4} & 0.81 & \textbf{0.62} \\
& \lightblue{\textbf{\system-3B (Ours)}} & \lightblue{200} & \lightblue{3.1B} & \lightblue{384} &
\lightblue{\textbf{1.89}} & \lightblue{\textbf{5.08}} & \lightblue{270.8} & \lightblue{\textbf{0.83}} & \lightblue{0.59} \\
\bottomrule
\end{tabular*}
\label{tab:imagenet_gen}
\end{table*}

\begin{figure}[bt]
\begin{center}
\centerline{\includegraphics[width=\columnwidth]{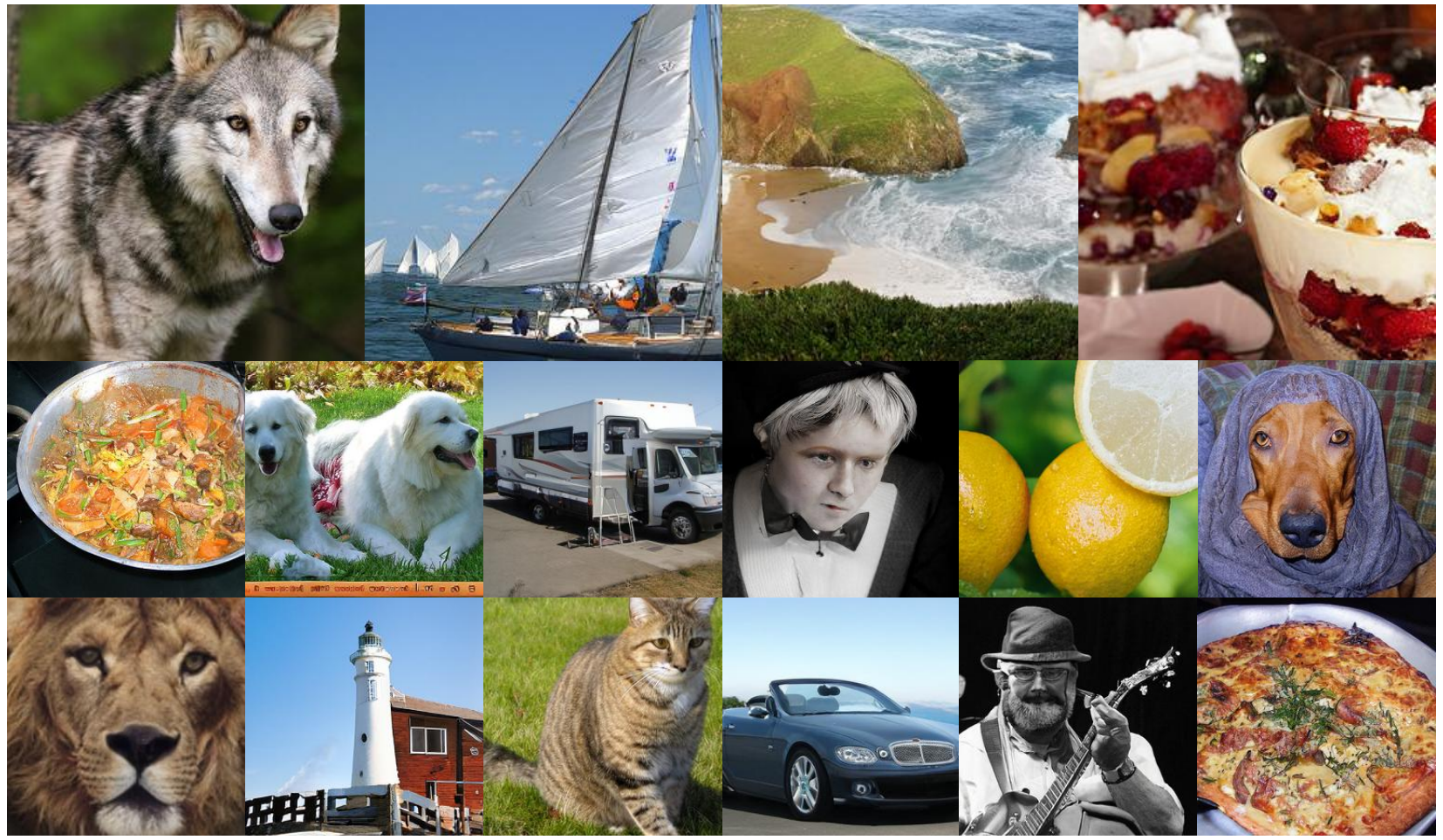}}
\caption{
Visualization of class-conditional image generation results from \system-L. 
}
\label{Fig4}
\end{center}
\end{figure}

\begin{table*}[t]
\centering
\small
\setlength{\tabcolsep}{1pt}
\renewcommand{\arraystretch}{1.25}
\caption{Comparison of tokenizer performance and AR generation at $256{\times}256$ resolution.}
\begin{tabular*}{\textwidth}{@{\extracolsep{\fill}}>{\raggedright\arraybackslash}p{0.18\textwidth}*{8}{>{\centering\arraybackslash}p{0.09\textwidth}}}
\hline
\multirow{2}{*}{Approach} & \multicolumn{3}{c}{Image recon.} & \multirow{2}{*}{Usage$\uparrow$} & \multirow{2}{*}{\#Epochs} & \multirow{2}{*}{\#Params.} & \multicolumn{2}{c}{AR gen.} \\
\cline{2-4} \cline{8-9}
 & \#Toks & rFID$\downarrow$ & rIS$\uparrow$ & & & & gFID$\downarrow$ & gIS$\uparrow$ \\
\midrule
LlamaGen-B & 256 & 2.22 & 169.8 & 95.2\% & 300 & 111M & 5.46 & 193.6 \\
VFMTok-B & 256 & 1.02 & 213.2 & 100.0\% & 300 & 111M & 3.61 & 247.6 \\
\system-B & 256 & 0.98 & 220.0 & 100.0\% & 300 & 111M & 3.43 & 181.9 \\
\bottomrule
\end{tabular*}
\label{tab:tokenizer_comparison_256}
\end{table*}

\begin{table*}[!h]
\centering
\small
\setlength{\tabcolsep}{6pt}
\renewcommand{\arraystretch}{1.15}
\caption{Ablations of \textbf{\system} along three axes: (a) fusion operator choices, (b) the effect of enabling spatial reconstruction, and (c) the backbone VFM. We test SigLIPv2~\cite{siglip2}, DINOv2~\cite{dinov2}, and DINOv3~\cite{dinov3} as VFM backbones. We report rFID as a measure of reconstruction fidelity and rIS as a measure of reconstruction semantic quality.}
\newcommand{\TABH}{1.2cm}
\begin{subtable}[t]{0.325\textwidth}
\centering

\begin{minipage}[t][\TABH][t]{\linewidth}
\centering
\begin{tabular}{lcc}
\toprule
Fusion type & rFID$\downarrow$ & rIS$\uparrow$ \\
\midrule
Attention & \textbf{0.61} & 230.4 \\
Linear    & 0.63          & 225.9 \\
None      & 0.85          & \textbf{231.1} \\
\bottomrule
\end{tabular}
\end{minipage}
\vspace{0.2in}
\caption{Fusion operator.}
\end{subtable}
\hfill
\begin{subtable}[t]{0.325\textwidth}
\centering

\begin{minipage}[t][\TABH][t]{\linewidth}
\centering
\begin{tabular}{lcc}
\toprule
Variant & rFID$\downarrow$ & rIS$\uparrow$ \\
\midrule
w/ $\mathcal{L}_{\mathrm{shallow}}$  & \textbf{0.61} & \textbf{230.4} \\
w/o $\mathcal{L}_{\mathrm{shallow}}$ & 0.66      & 229.4 \\
\bottomrule
\end{tabular}
\end{minipage}
\vspace{0.16in}
\caption{Shallow alignment.}
\end{subtable}
\hfill
\begin{subtable}[t]{0.325\textwidth}
\centering

\begin{minipage}[t][\TABH][t]{\linewidth}
\centering
\begin{tabular}{lcc}
\toprule
VFM variant & rFID$\downarrow$ & rIS$\uparrow$ \\
\midrule
SigLIP2 & 0.61 & \textbf{230.4} \\
DINOv2   & 0.60 & 227.0 \\
DINOv3   & \textbf{0.54}   & 227.9 \\
\bottomrule
\end{tabular}
\end{minipage}
\vspace{0.2in}
\caption{Backbone VFM.}
\end{subtable}

\label{tab:ablation_main}
\end{table*}

\paragraph{Class-conditional image generation.}
We compare ag\-ainst representative mainstream generators, including diffusion models (Diff.)~\cite{dit,sit,fasterdit,maskdit}, masked generation models (Mask.)~\cite{maskgit}, and autoregressive models (AR) built on visual tokenizers or semantic tokenizers~\cite{llamagen,var,titok,vfmtok}.
All AR baselines are trained and evaluated under the same protocol as LlamaGen~\cite{llamagen}.

As shown in Table~\ref{tab:imagenet_gen}, \system yields strong generation performance compared to mainstream image generation models.
Notably, at the Base scale, \system-B achieves a gFID of $3.38$, outperforming masked autoregressive baselines~\cite{maskgit} with fewer parameters. \system-B also substantially outperforms AR baselines trained on visual tokenizers such as LlamaGen~\cite{llamagen}, with a gain of $2.71$ in gFID and a gain of $37.3$ in gIS.
When scaled to the Large scale, \system-L further reduces gFID to 2.26, which is comparable to some competitive diffusion models~\cite{maskdit,dit,sit,fasterdit}. However, \system requires much shorter training length  and approximately half of the parameters needed by diffusion models, demonstrating the efficiency of our model.

Scaling \system to larger models further improves generation quality.
At the XXL scale, \system-XXL reaches a gFID of 1.95 and the best sFID of 4.81, surpassing strong AR baselines such as VFMTok-XXL~\cite{vfmtok} and LlamaGen-XXL~\cite{llamagen} under the same training length.
Notably, when scaled to 3B parameters, \system continues improving and achieves a gFID of \textbf{1.89}, establishing a new state-of-the-art result for autoregressive modeling.

VFMTok~\cite{vfmtok} achieves higher gIS than \system at similar parameter counts. We attribute this difference to a well-known trade-off between IS and FID. IS emphasizes classification confidence, which does not necessarily reflect the image realism captured by FID.
Moreover, \system has tighter training constraints: its decoded feature must stay close to the underlying VFM semantic geometry while remain directly decodable by a CNN pixel head for high-fidelity reconstruction.
These additional constraints reduce the degrees of freedom available for generation-optimality, which can manifest as lower gIS even when fidelity-oriented metrics (\eg, gFID/sFID) remain strong.

Overall, these results show that \system provides a three-in-one unified representation, supporting AR modeling without sacrificing VFM semantics during semantic tokenization.

\subsection{Ablation Study}
\label{sec:ablation}

We conduct ablations from two complementary perspectives. First, we provide a controlled comparison at $256{\times}256$ resolution to isolate the effect of the tokenizer. Then, we analyze the core design choices of \system, including feature fusion, shallow-feature supervision, and the choice of VFM backbone.

\paragraph{Controlled $256{\times}256$ AR Generation.}
Following VFMTok, we train both the image tokenizer and the AR generation model at $256{\times}256$ resolution. The tokenizer is trained for 50 epochs, and the AR-Base model is trained for 300 epochs. As shown in \cref{tab:tokenizer_comparison_256}, \system improves over LlamaGen and VFMTok in both reconstruction quality and generation fidelity under this controlled setting.

\paragraph{Design Ablations.}
\cref{tab:ablation_main} examines three design aspects of \system: the fusion operator, the auxiliary supervision for shallow spatial reconstruction, and the choice of VFM backbone.

We first find that fusion is critical: removing fusion leads to a clear drop in reconstruction quality, confirming that injecting complementary shallow spatial cues into deep semantic features is essential for decodability.
Across fusion choices, reconstruction fidelity is relatively stable, while semantic retention is more sensitive: in particular, attention better preserves semantics under reconstruction-driven learning, suggesting it is more effective at selectively integrating low-level details without distorting the semantic structure.

Next, adding the auxiliary objective to reconstruct shallow features consistently improves reconstruction, validating the benefit of explicitly supervising reconstruction-friendly signals in the decoder.

Finally, \system is robust across different VFM backbones, achieving strong performance trends consistently.
We observe a mild trade-off between reconstruction and semantics: DINO-style~\cite{dinov2,dinov3} SSL features tend to favor reconstruction, whereas SigLIP2~\cite{siglip2} features better support semantic retention and offer vision--language aligned representations that can directly interact with text.
For this reason, we adopt SigLIP2 as the default backbone in our main experiments.

\section{Conclusion}

We introduced \textbf{\system}, a discrete representation autoencoder that converts VFM features into discrete codes for autoregressive image generation while preserving both semantic richness and high-fidelity reconstructability. The design is motivated by a simple empirical observation: VFM feature hierarchies exhibit a clear depth-dependent trade-off, where shallow layers retain spatial detail useful for reconstruction, whereas deeper layers encode stronger semantics. Exploiting this complementarity, \system injects reconstruction-relevant shallow signals into deep semantic features, yielding a latent space that retains both detailed visual information and strong semantics. Experiments on ImageNet show that \system delivers strong performance on both reconstruction and generation, while largely preserving the semantics of the original VFM representations. When scaled to 3B parameters, \system achieves a gFID of $\mathbf{1.89}$ at $256\times256$, establishing a new state of the art for autoregressive image generation. 

\bibliographystyle{plainnat}
\bibliography{main}

\clearpage
\appendix

\section{\system Implementation Details}
\label{appendix:settup}
\subsection{Tokenizer Training Details}
Overall, our tokenizer training recipe closely follows prior work VFMTok~\cite{vfmtok}.
Since VFMTok uses a VFM with a patch size of 14 and an input resolution of 336, we use a patch size of 16 and an input resolution of 384 to maintain consistency in the feature map size. We train \system on ImageNet-1K~\cite{imagenet} training set using random resized crop and horizontal flip, with an input resolution of $384\times384$ and evaluating reconstructions at $256\times256$ following the common protocol~\cite{llamagen}. \system requires 2 days of training on 8 Nvidia H200 GPUs. We summarize some key training configuration of our tokenizer in Table~\ref{tab:tokenizer_detail}.

\begin{table}[htbp]
\centering
\caption{Tokenizer implementation details}
\label{tab:impl_tokenizer}
\setlength{\tabcolsep}{6pt}
\renewcommand{\arraystretch}{1.15}

\begin{tabular*}{\columnwidth}{@{\extracolsep{\fill}} l l l l @{}}
\toprule
Hyperparameter & Value & Hyperparameter & Value \\
\midrule

\multicolumn{4}{@{}l}{\textbf{Backbone}} \\
VFM type & SigLIPv2-Large~~\cite{siglip2}
& VFM input resolution & 384 \\
VFM training & Frozen
& image\_size / eval\_image\_size & 384 / 256 \\
decoder backbone & ViT
& decoder layer\_num & 6 \\
decoder hidden\_dim & 1024
& decoder attn\_head & 8 \\
decoder cls\_num & 1
& decoder reg\_num & 4 \\
decoder dropout & 0.1 \\
\midrule

\multicolumn{4}{@{}l}{\textbf{General}} \\
mixed\_precision & bf16
& ema & True \\
codebook\_l2\_norm & True
& max\_grad\_norm & 1.0 \\
\midrule

\multicolumn{4}{@{}l}{\textbf{Loss}} \\
reconstruction\_weight & 1.0
& perceptual\_weight & 1.0 \\
vq\_loss\_ratio & 1.0
& commit\_loss\_beta & 0.25 \\
\midrule

\multicolumn{4}{@{}l}{\textbf{Adversarial}} \\
disc\_type & dino
& disc\_loss / gen\_loss & hinge / hinge \\
disc\_weight & 0.5
& disc\_start & 20000 \\
use\_diff\_aug & True
&  &  \\
\midrule

\multicolumn{4}{@{}l}{\textbf{Optimization}} \\
epochs & 50
& global\_batch\_size & 256 \\
optimizer & AdamW~~\cite{adamW}
& lr / lr\_scheduler & 1e-4 / cosine \\
weight\_decay & 5e-2
& beta1 / beta2 & 0.9 / 0.95 \\
\bottomrule
\end{tabular*}
\label{tab:tokenizer_detail}
\end{table}

\subsection{Autoregressive Training Details}
Following VFMTok~\cite{vfmtok}, an AR generator is trained to model the discrete token sequences produced by the tokenizer.
However, VFMTok extracts token sequences on-the-fly using the tokenizer at each training epoch, which introduces additional overhead.
In contrast, we follow the original LlamaGen training pipeline~\cite{llamagen}: we apply ten-crop preprocessing to training images and pre-extract all token sequences offline before AR training, significantly improving training throughput.
We train Base and Large models for 300 epochs, and train XXL and 3B models for 200 epochs, consistent with the scaling recipe in VFMTok.
\system-B takes approximately 34 hours of training on 8 Nvidia H200 GPUs. Key AR training hyperparameters are summarized in Table~\ref{tab:ar_detail}.

\begin{table}[t]
\centering
\small
\caption{Autoregressive training and sampling configuration for \system.}
\label{tab:ar_detail}
\setlength{\tabcolsep}{3pt}
\renewcommand{\arraystretch}{1.15}

\begin{tabular*}{\columnwidth}{@{\extracolsep{\fill}} l l l l @{}}
\toprule
Hyperparameter & Value & Hyperparameter & Value \\
\midrule

\multicolumn{4}{@{}l}{\textbf{Training protocol}} \\
token extraction & offline
& image preprocessing & ten-crop \\
epochs (Base/Large) & 300
& epochs (XXL/3B) & 200 \\
EMA & True
& mixed\_precision & bf16 \\
\midrule

\multicolumn{4}{@{}l}{\textbf{Optimization}} \\
optimizer & AdamW~\cite{adamW}
& lr & 1e-4 \\
weight\_decay & 0.05
& beta1 / beta2 & 0.9 / 0.95 \\
max\_grad\_norm & 1.0
& dropout\_p & 0.1 \\
token\_dropout\_p & 0.1
& drop\_path\_rate & 0.0 \\
\midrule

\multicolumn{4}{@{}l}{\textbf{Architecture \& conditioning}} \\
class\_token\_num & 1
& class\_dropout\_prob & 0.1 \\
positional embedding & 2D RoPE~\cite{rope}
& rope\_base & 10000 \\
\midrule

\multicolumn{4}{@{}l}{\textbf{Sampling}} \\
top\_k & 0
& top\_p & 1.0 \\
temperature & 1.0
&  &  \\
\bottomrule
\end{tabular*}
\end{table}

\section{Additional Qualitative Results}
We provide additional qualitative results on image reconstruction and generation, and further analyze representative failure cases of both our tokenizer and autoregressive model.
\subsection{Reconstruction Results}
\begin{figure*}[t]
\vskip 0.2in
\begin{center}
\centerline{\includegraphics[width=\textwidth]{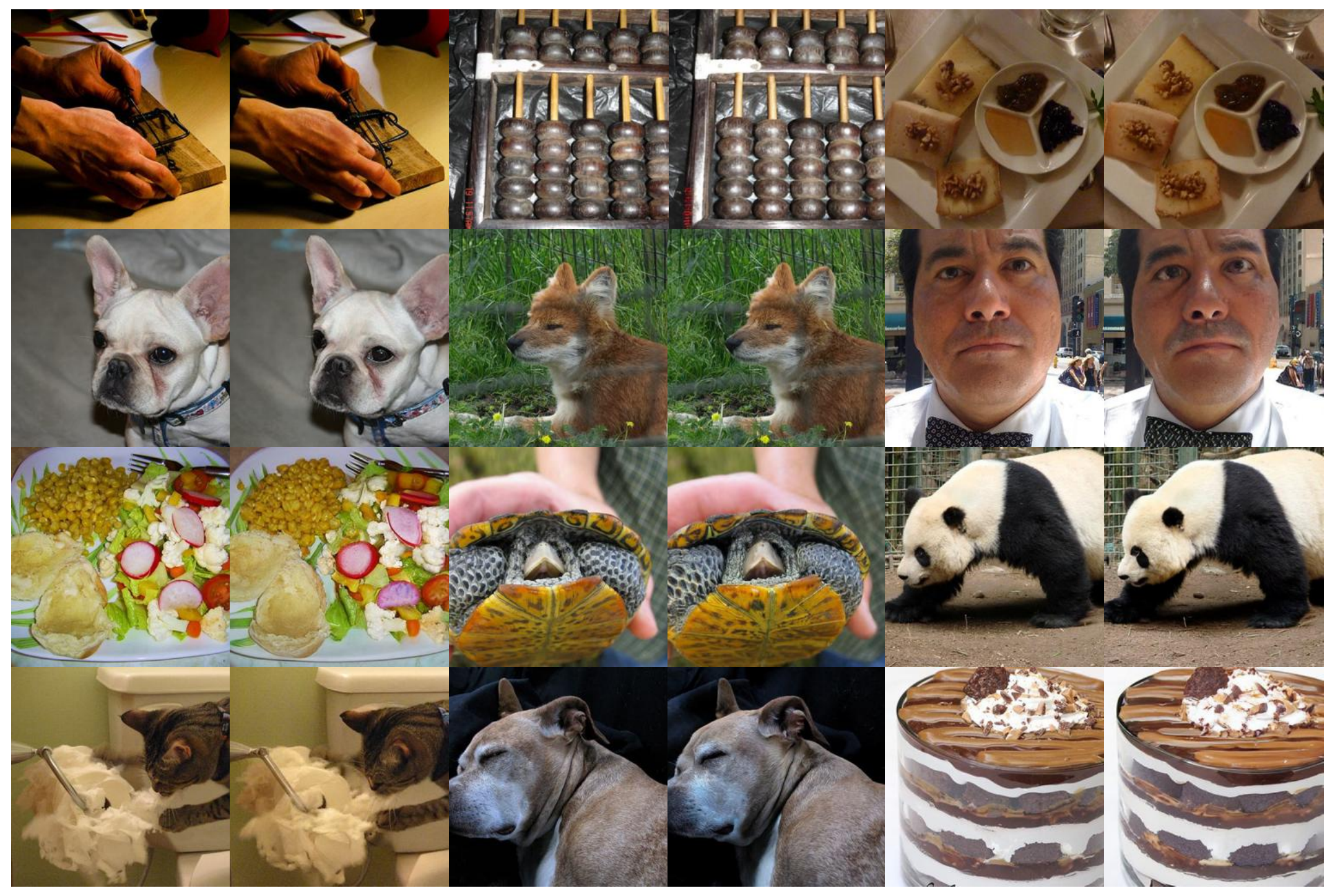}}
\caption{
More visualization of reconstruction results from \system. Left: input image; Right: output image.
}
\label{Fig5}
\end{center}
\end{figure*}

As shown in Figure~\ref{Fig5}, our tokenizer produces fine-grained reconstructions across diverse scenes and objects.
\subsection{Generation Results}
\begin{figure*}[t]
\vskip 0.2in
\begin{center}
\centerline{\includegraphics[width=\textwidth]{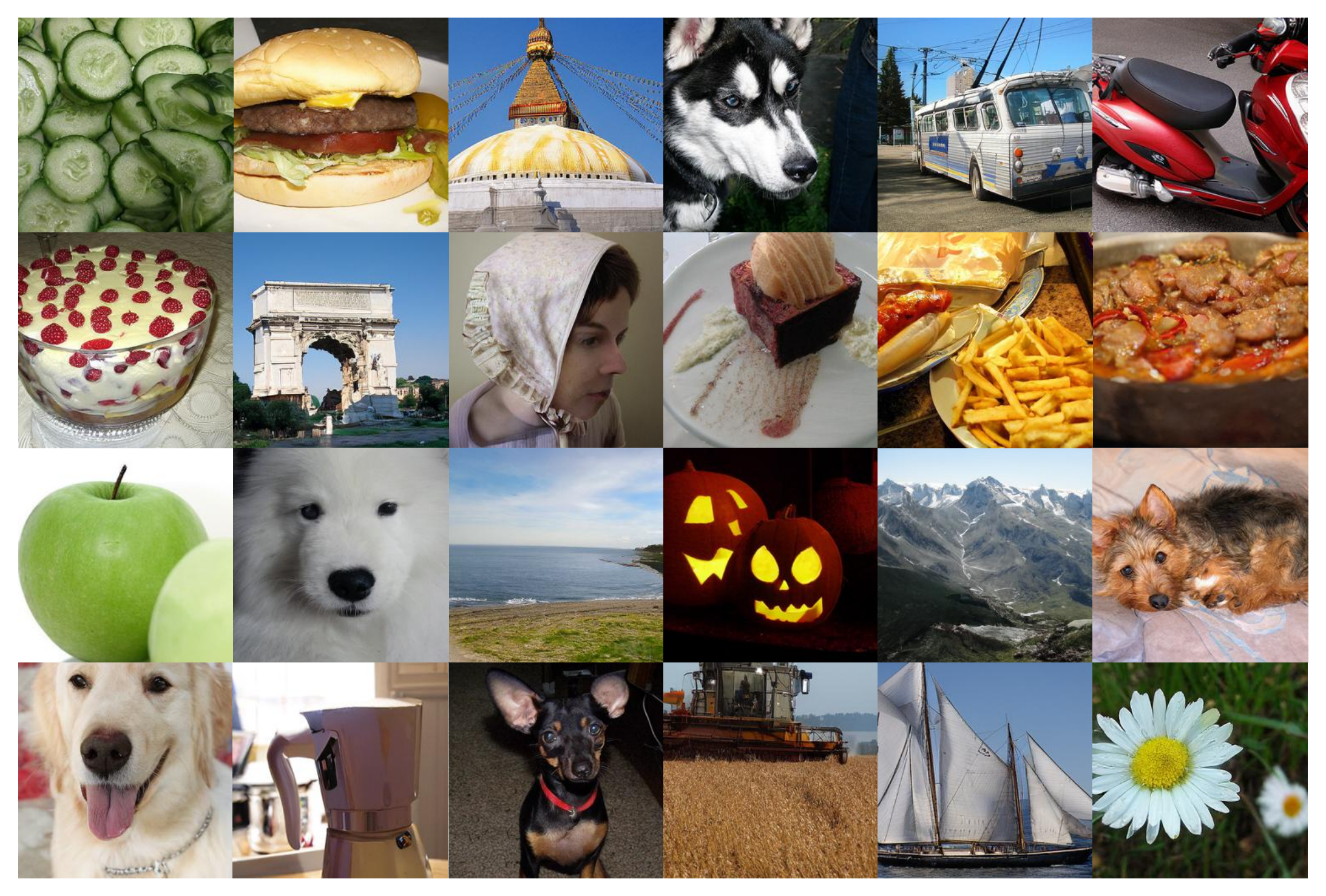}}
\caption{
More visualization of class-conditional image generation results from \system-L. 
}

\label{Fig6}
\end{center}
\end{figure*}

Figure~\ref{Fig6} presents more samples demonstrating that our method can synthesize images with varied styles, subjects, and compositions.
\subsection{Failure Cases}
\begin{figure*}[t]
\vskip 0.2in
\begin{center}
\centerline{\includegraphics[width=\textwidth]{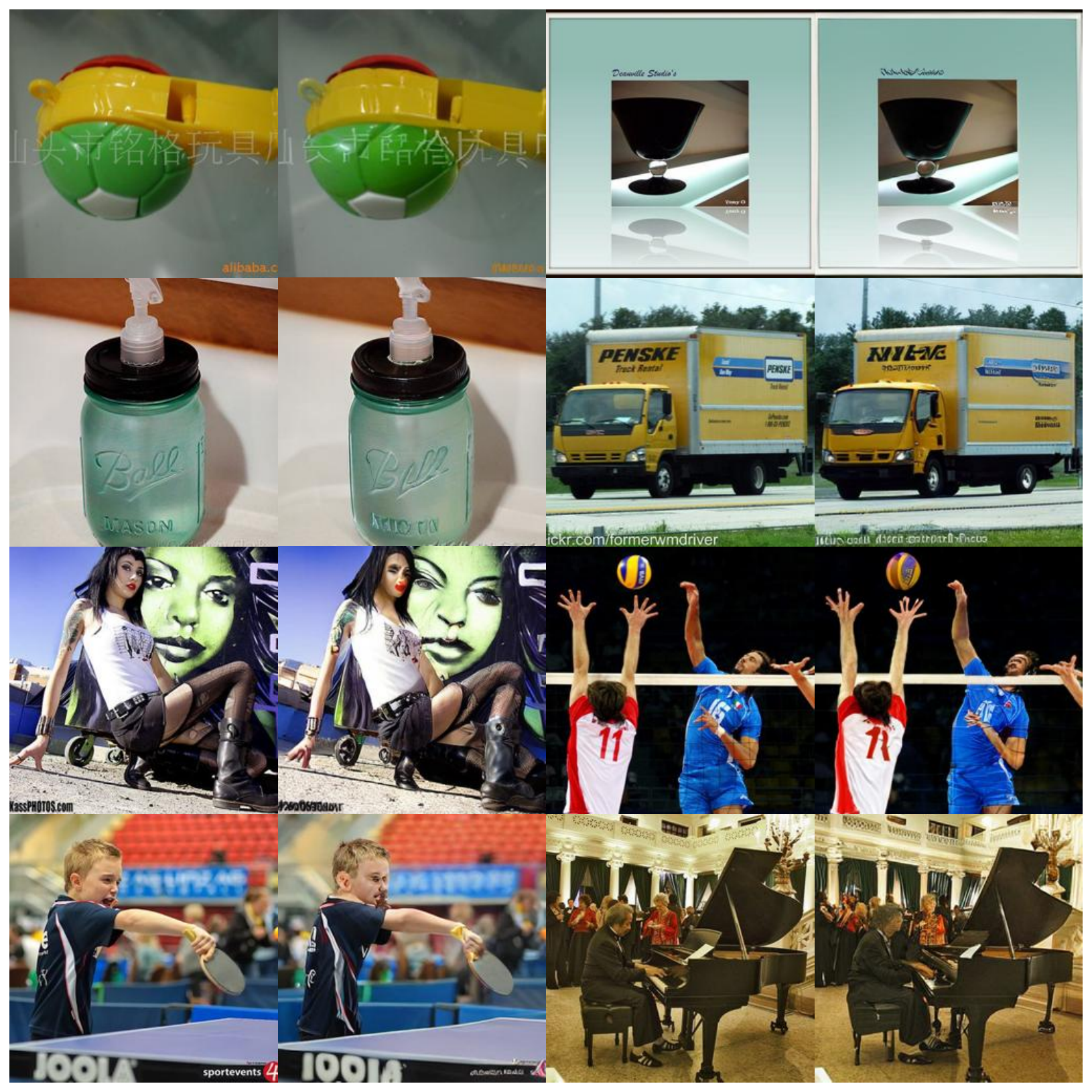}}
\caption{
Visualization of failure reconstruction cases from \system. Left: input image; Right: output image. 
}

\label{Fig7}
\end{center}
\end{figure*}

\begin{figure*}[t]
\vskip 0.2in
\begin{center}
\centerline{\includegraphics[width=\textwidth]{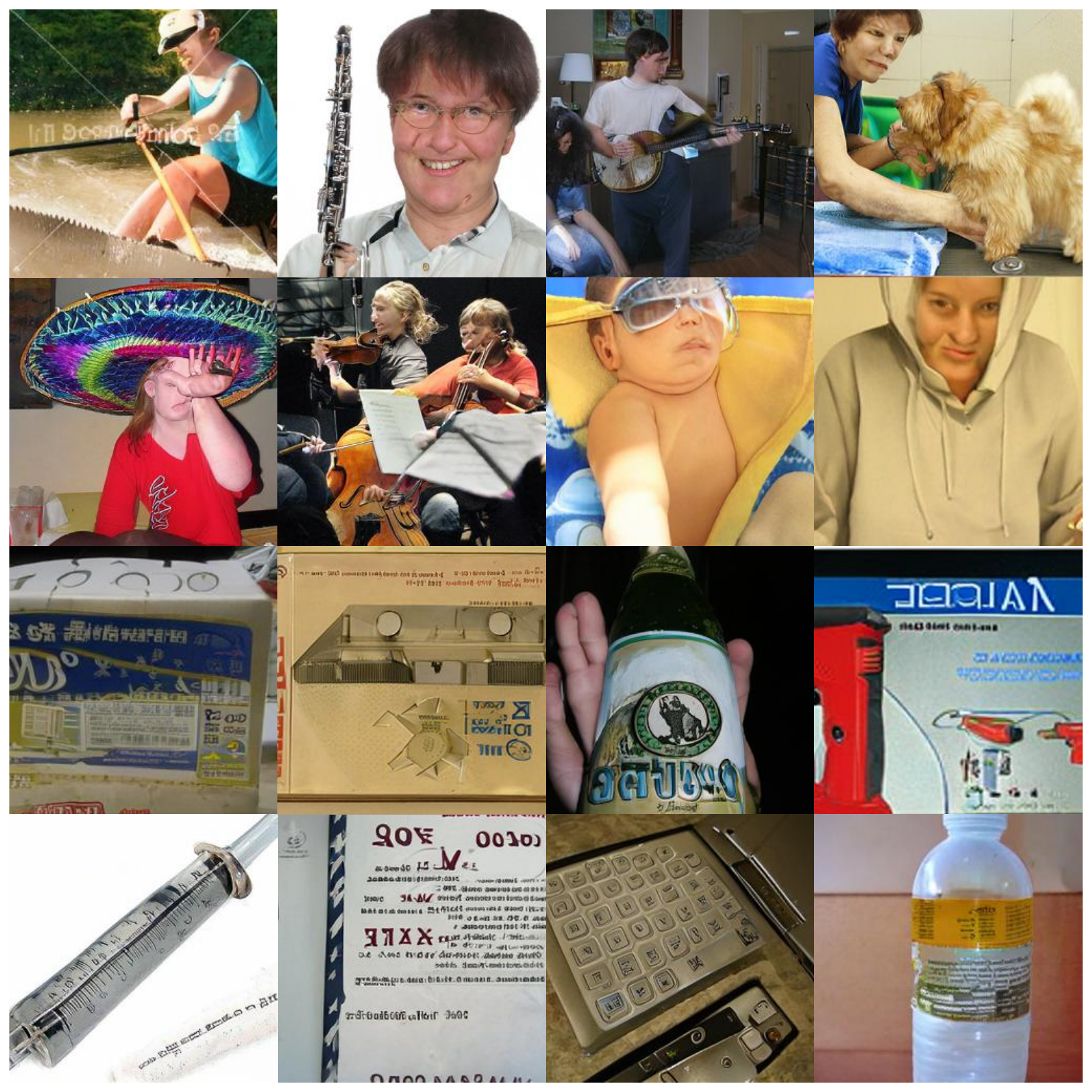}}
\caption{
Failure generation cases. \system still has artifacts in generating delicate text, human faces and fingers, which can be addressed with more training data on these images. 
}

\label{Fig8}
\end{center}
\end{figure*}

Despite these strengths, we observe degraded reconstruction quality on faces and text, as illustrated in Figure~\ref{Fig7}. We attribute this to the limited domain coverage of our tokenizer training data. In particular, our tokenizer is trained only on ImageNet, which contains sparse coverage of close-up faces and rich-text images. We do not incorporate additional face- or text-centric data either. In generation, Figure~\ref{Fig8} shows that artifacts can still appear in fine-structure regions such as hands and faces, suggesting that post-training refinement on autoregressive models may be beneficial for further improving fidelity.

\section{Limitation and Future Work}
\label{sec:limit}
Our tokenizer is trained mainly on ImageNet, which has limited domain coverage. Thus, reconstruction can degrade on faces, text, and other long-tail visual patterns.
In addition, our semantic-preservation evaluation focuses on ImageNet zero-shot classification, which mainly reflects category-level semantics and does not fully cover broader semantic capabilities.

A direct next step is to pretrain or adapt the tokenizer on larger and more diverse datasets to improve coverage of faces, text, and long-tail domains.
We also plan to evaluate the decoded interface feature on broader semantic benchmarks to better characterize semantic preservation~\cite{pope,gqa,textvqa}.
Finally, our discrete-token formulation may naturally extend to videos by incorporating temporal consistency for semantic tokenization and generation.

\end{document}